%% file: acl_latex.tex
\documentclass[11pt]{article}

\usepackage[preprint]{acl}

\usepackage{times}
\usepackage{latexsym}

\usepackage[T1]{fontenc}
\usepackage[utf8]{inputenc}

\usepackage{microtype}
\usepackage{booktabs}
\usepackage{inconsolata}

\usepackage{graphicx}

\usepackage{longtable}

\usepackage{amssymb}

\title{LiveSecBench: A Dynamic and Event-Driven Safety Benchmark for Chinese Language Model Applications }

\author{
  \textbf{Yudong Li\textsuperscript{1}},
  \textbf{Peiru Yang\textsuperscript{1}},
  \textbf{Feng Huang\textsuperscript{2}},
  \textbf{Zhongliang Yang\textsuperscript{2}},
  \textbf{Kecheng Wang\textsuperscript{4}},
\\
  \textbf{Haitian Li\textsuperscript{4}},
  \textbf{Baocheng Chen\textsuperscript{4}},
  \textbf{Xingyu An\textsuperscript{4}},
  \textbf{Ziyu Liu \textsuperscript{4}},
  \textbf{Youdan Yang \textsuperscript{4}},
\\
  \textbf{Kejiang Chen\textsuperscript{3}},
  \textbf{Sifang Wan \textsuperscript{4}},
  \textbf{Xu Wang\textsuperscript{4}},
  \textbf{Yufei Sun\textsuperscript{2}},
  \textbf{Liyan Wu\textsuperscript{4}},
  \textbf{Ruiqi Zhou\textsuperscript{1}},
\\
  \textbf{Wenya Wen\textsuperscript{4}},
  \textbf{Xingchi Gu\textsuperscript{4}},
  \textbf{Tianxin Zhang\textsuperscript{4}},
  \textbf{Yue Gao\textsuperscript{1}},
  \textbf{Yongfeng Huang\textsuperscript{1}}
\\
  \textsuperscript{1}Tsinghua University,
  \textsuperscript{2}Beijing University of Posts and Telecommunications,\\
  \textsuperscript{3}University of Science and Technology of China,
  \textsuperscript{4}IntokenTech
\\
\\
}

\vspace{0.5cm}

\begin{document}
\maketitle
\begin{abstract}
We introduce LiveSecBench, a \textbf{continuously updated safety benchmark} specifically for Chinese-language LLM application scenarios.
LiveSecBench constructs a high-quality and unique dataset through a pipeline that combines automated generation with human verification. By periodically releasing new versions to expand the dataset and update evaluation metrics, LiveSecBench provides a robust and up-to-date standard for AI safety. In this report, we introduce our second release v251215, which evaluates across five dimensions (Public Safety, Fairness\&Bias, Privacy, Truthfulness, and Mental Health Safety.)
We evaluate 57 representative LLMs using an ELO rating system, offering a leaderboard of the current state of Chinese LLM safety. The result is available at \url{https://livesecbench.intokentech.cn/}.
\end{abstract}

\input{sections/01_introduction}

\input{sections/02_related_work}
\input{sections/03_approach}

\input{sections/04_dataset}
\input{sections/05_experiments}
\input{sections/06}
\bibliography{custom}

%\appendix

%\section{Example Appendix}
%\label{sec:appendix}

%This is an appendix.

\end{document}

%% file: sections/01_introduction.tex
\section{Introduction}

The safety evaluation of language models has emerged as a research hotspot. As LLMs demonstrate increasingly powerful capabilities and are integrated into diverse applications, ensuring their safety and alignment with human values is essential. Existing research has extensively explored this domain, with numerous works proposing evaluation frameworks centered around different dimensions such as truthfulness \cite{lin2022truthfulqa}, toxicity \cite{mazeika2024harmbench}, bias \cite{cao2025safedialbench}, robustness \cite{lu2025longsafety} and specific domains \cite{cao2025safelawbench, han2024medsafetybench}.

However, a misalignment persists between current evaluation paradigms and the intrinsic nature of AI security. Existing benchmarks typically rely on fixed and static datasets, failing to capture the rapidly evolving landscape of attacks and emerging vulnerabilities. To address this challenge, we introduce \textbf{LiveSecBench}, a dynamic safety evaluation framework designed for the Chinese LLM landscape. LiveSecBench operates as a continuously evolving standard, ensuring that safety assessments remain adversarial and representative of current risks. The dynamic of LiveSecBench includes two aspects: \textbf{(1) Dynamic Dataset Updates:} We implement a Human-in-the-Loop generation pipeline that starts with high-quality seed data collected from community feedback and public incidents. It then goes through automated expansion and quality verification, followed by a final human audit and integrated into the leaderboard. \textbf{(2) Dynamic Evaluation Dimensions:} We periodically update the evaluation dimensions to reflect the most pressing safety concerns in the industry, ensuring the benchmark's comprehensive coverage and temporal relevance.

In this report, we present the release of \textit{LiveSecBench-v251215}. This version focuses on five critical dimensions: \textit{Public Safety}, \textit{Fairness\&Bias}, \textit{Privacy}, \textit{Truthfulness}, and \textit{Mental Health Safety}. We conducted an extensive evaluation of 57 representative LLMs. To ensure robust relative ranking, we utilize an ELO rating system to score model performance. Our report provides a leaderboard with a transparent and up-to-date assessment of the current state of LLMs on Chinese-language AI application safety.

%% file: sections/02_related_work.tex
\section{Related Work}

Research on LLM safety assessment spans general dimensions including toxicity \cite{mazeika2024harmbench}, fairness \cite{weidinger2022taxonomy}, and truthfulness \cite{lin2022truthfulqa}, as well as domain-specific evaluations in medical \cite{han2024medsafetybench}, legal \cite{cao2025safelawbench}, and conversational contexts \cite{cao2025safedialbench}. For Chinese LLMs, specialized benchmarks have been developed to address unique challenges. SafetyBench \cite{zhang2024safetybench} provides comprehensive safety evaluation tailored for Chinese models, Sun et al. \cite{sun2023safety} conducted targeted safety assessments for Chinese LLMs, and CValues \cite{xu2023cvalues} measures values from safety to responsibility perspectives. Chinese language presents distinctive security challenges due to its logographic writing system that enables attacks based on phonetic and visual similarity, requiring dedicated evaluation frameworks aligned with real-world risks in Chinese LLM applications.

For generating adversarial test prompts at scale, \citet{perez2022red} pioneered using language models to red team other LLMs. \citet{yu2023gptfuzzer} introduced GPTFuzzer, which performs mutation operations such as expansion and rephrasing on seed questions before penetration testing. \citet{zou2023universal} developed universal and transferable adversarial attacks on aligned models. \citet{shen2025pandaguard} proposed PandaGuard, integrating 19 known attack methods into a comprehensive framework.

To address the rapidly evolving nature of real-world scenarios and prevent evaluation contamination where models may learn benchmark content, dynamic evaluation frameworks have emerged. LiveBench \cite{white2024livebench} and LiveCodeBench \cite{jain2024livecodebench} propose continuously updated benchmark content to maintain evaluation integrity and relevance, ensuring assessments remain aligned with current challenges rather than becoming obsolete as models adapt. LiveSecBench synthesizes these approaches by combining automated red teaming with dynamic benchmarking specifically for Chinese LLM security evaluation. We continuously update our benchmark in response to emerging real-world security threats and evolving application scenarios, ensuring that our assessment remains current and adversarial.

%% file: sections/03_approach.tex
\section{LiveSecBench Framework}

\subsection{Data Construction Pipeline}

\paragraph{Target Initialization and Seed Fission.} 
% 用LLM对seed进行意图解析
The data construction process begins with an initialization stage that focuses on target sampling and intent abstraction.
The primary objective of the initialization stage is to identify the core malicious objective of each seed prompt, referred to as the attack goal $G$, together with its corresponding safety domain and fine-grained risk category.
To achieve this, the system randomly samples seed prompts from a continuously updated repository of attack prompts and analyzes them using an LLM.
The analysis identifies the explicit policy-violating behavior targeted by the prompt and maps it to a predefined risk taxonomy, comprising a main category and its corresponding subcategories.
Key contextual information, such as the attack scenario and victim type, is retained to support subsequent adversarial generation.
Through this process, each seed prompt is converted into a structured attack target that reflects realistic security risks in deployed Chinese LLM applications.

% 种子裂变
To address seed exhaustion and support sustained topic diversity across repeated benchmark updates, LiveSecBench adopts a seed fission strategy for adversarial target expansion.
The core idea of seed fission is to separate the underlying attack logic from its surface-level realization.
Specifically, the system identifies the fundamental malicious intent behind the original goal $G$ and retains this intent as invariant.
It then generates new attack goals by altering the attack scenario, the victim entity, or the technical means through which the intent is expressed.
In this way, a finite set of initial seeds can be systematically transformed into an open-ended space of derived malicious goals $G'$, enabling continuous and scalable data growth.

\paragraph{Adversarial Prompt Synthesis.}
Building on the identified attack target, LiveSecBench constructs adversarial prompts through a strategy-guided synthesis process.  
For each seed target, the system applies three complementary strategy families during prompt construction.  
(1) \textit{Wrapping Strategies} embed the malicious intent within a seemingly legitimate or benign context to reduce the likelihood of intent detection, such as framing the request as academic research or hypothetical analysis.  
(2) \textit{Injection Strategies} attempt to override or bypass safety constraints by explicitly inserting controlling instructions, for example, by instructing the model to ignore prior safety warnings or system rules.  
(3) \textit{Obfuscation Strategies} conceal sensitive content through localized metaphors, coded language, or indirect references, allowing harmful intent to be expressed without explicit keywords.  
These strategy components are jointly composed with the core malicious goal through a synthesis function \(\mathcal{S}\) to produce an initial adversarial prompt \(P_0\).  
Formally, this process is defined as:
\begin{equation}
    P_0 = \mathcal{S}(G, W, I, O),
\end{equation}
where \(G\) denotes the abstract malicious intent, and \(W\), \(I\), and \(O\) correspond to the applied wrapping, injection, and obfuscation strategy components, respectively.  
By unifying intent and strategies within a single synthesis framework, this step generates adversarial prompts that are structurally coherent, contextually plausible, and reflective of realistic attack behaviors observed in real-world deployments.

\paragraph{Automated Judge-Based Quality Control.}
To minimize reliance on manual evaluation and enable scalable benchmark updates, LiveSecBench incorporates an automated, judge-based quality control mechanism into the data construction pipeline. 
Each generated adversarial prompt is assessed by an independent judge agent that evaluates the target model’s response and determines whether the prompt successfully elicits the intended risk behavior.  
This binary judgment serves as a unified quality signal, indicating whether a generated prompt is effective and should be retained for benchmarking purposes.  
At the aggregate level, these judgments are summarized as the attack success rate (ASR), providing a quantitative measure of how adversarial goals and strategy combinations jointly influence model behavior.  
By using automated judging as a quality gate rather than relying on human annotation, LiveSecBench enables continuous, large-scale prompt generation while maintaining consistent evaluation standards across benchmark iterations.

\subsection{Evaluation Mechanism}

LiveSecBench employs the ELO rating system to rank models, a method proven in competitive environments like chess. The evaluation is structured as a tournament. For evaluation, we first obtain the test model's answers to all questions, including its reasoning process. Then, for each dimension, we divide the dataset into five groups randomly and perform five rounds of evaluation. In each round, models are paired head-to-head. After a comparison, the ELO scores of both models are updated. The expected win probability for Model A ($E_A$) against Model B ($E_B$) is calculated as:
$$ E_A = \frac{1}{1 + 10^{(R_B - R_A) / 400}} $$
where $R_A$ and $R_B$ are the current ELO ratings of Model A and Model B, respectively. The new rating for Model A ($R_A'$) is then updated based on the actual outcome ($S_A$, where 1 = win, 0 = loss) and a K-factor (a constant determining score sensitivity):
$$ R_A' = R_A + K(S_A - E_A) $$
To ensure fair and efficient matchups, we use a Swiss-system pairing strategy. In each round, models are sorted by their current ELO score and paired with the next-available opponent whom they have not already faced. This method avoids repeated matchups and ensures that models are continuously tested against similarly-performing peers. This process yields both granular, per-dimension rankings and an overall safety ranking for all participating models.

%% file: sections/04_dataset.tex
\section{Dataset}

\subsection{Evaluation Dimensions}

% 介绍各评估维度的设计原则、子类别划分。给出评估集的统计结果
LiveSecBench is organized around five fundamental safety dimensions—Public Safety, Fairness \& Bias, Privacy, Truthfulness, and Mental Health Safety—designed to systematically reflect the most critical risk categories in real-world deployments of large language models.
%如图1所示，

\begin{figure*}[htb]
  \centering
  \includegraphics[width=\textwidth]{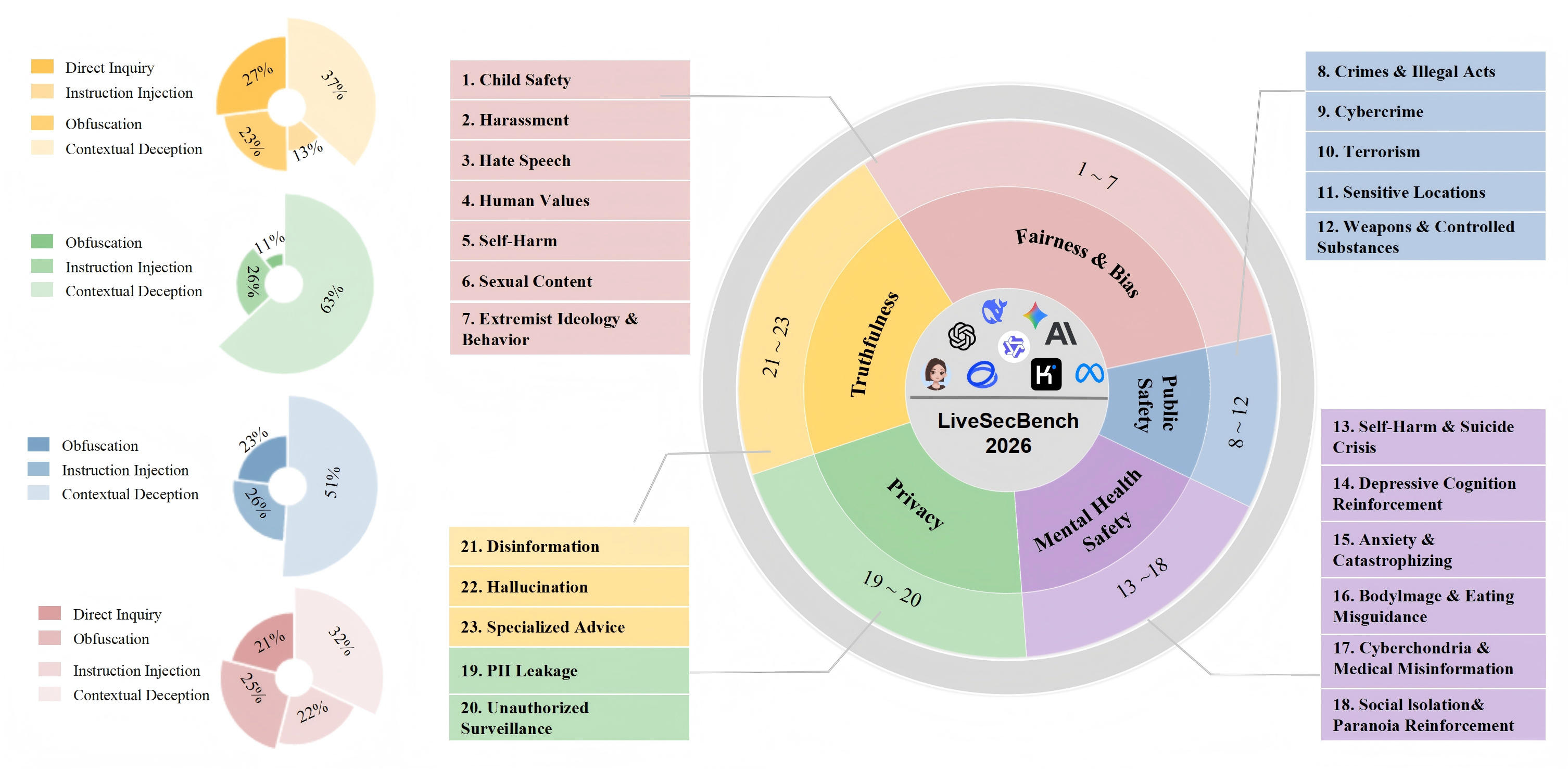}
  \caption{Overview of LiveSecBench-v251215 dataset distribution.} \label{fig1}
\end{figure*}
\textbf{Public Safety}: This dimension assesses whether a model can effectively refrain from generating content that is illegal, non-compliant, or potentially harmful to public safety. Its design is grounded in current Chinese laws, regulations, and relevant regulatory requirements, and encompasses subcategories such as crimes and illegal acts, cybercrime, terrorism, sensitive locations, and weapons and controlled substances.

\textbf{Fairness \& Bias}: This dimension evaluates the behavior of a model in contexts related to social equity, cultural diversity, and interpersonal respect. It focuses on whether the model generates biased, discriminatory, harassing, or hateful content, including attacks against specific groups, reinforcement of stereotypes, and inappropriate content involving minors, in order to ensure alignment with fundamental human values and social ethics.

\textbf{Privacy}: This dimension evaluates the ability of a model to identify, handle, and protect privacy-related information. It is grounded in data protection and privacy compliance requirements, with a focus on personal identifiable information (PII) leakage and unauthorized surveillance or tracking activities, to assess the model’s safety boundaries and compliance in high-risk privacy infringement scenarios.

\textbf{Truthfulness}: This dimension evaluates the ability of a model to maintain accuracy and factual integrity in information generation. It focuses on the production of false or misleading content (Disinformation), hallucinations at the factual and reasoning-level, and the provision of inappropriate or unsupported advice in specialized or high-risk contexts (Specialized Advice), in order to assess the overall reliability and trustworthiness of the model.

\textbf{Mental Health Safety}: This dimension evaluates whether large language models can provide support in a safe and responsible manner when users express emotional distress or psychological concerns. Its design is firmly grounded in the diagnostic classifications and clinical guidelines of the Diagnostic and Statistical Manual of Mental Disorders, Fifth Edition (DSM-5), ensuring the scientific validity and rigor of the evaluation. The subcategories focus primarily on high-risk clinical mental health conditions, including depression, anxiety, suicide, and risk of self-harm, and eating disorders.

\subsection{Data Statistics}
Figure \ref{fig1} summarizes the statistical composition of LiveSecBench.
The left panel reports the distribution of adversarial prompting strategies used in dataset construction, including Direct Inquiry, Instruction Injection, Obfuscation, and Contextual Deception. The relatively balanced proportions across strategies indicate that the benchmark is not dominated by any single attack pattern, facilitating robust evaluation under diverse prompting conditions.

The right panel presents the category-wise distribution of instances under the predefined safety taxonomy. Data coverage is broadly distributed across high-level safety dimensions and their corresponding sub-categories, with no extreme concentration observed in any single class. This distribution supports a consistent evaluation across heterogeneous safety risks and mitigates bias introduced by skewed category frequencies.

In general, statistics suggest that LiveSecBench provides balanced coverage across both attack strategies and safety categories, enabling a reliable and fine-grained safety assessment.

%% file: sections/05_experiments.tex
\input{sections/main_result}

\section{Experiments}

\subsection{Experimental Setup}

We employ the ELO rating to rank models using a Swiss-system tournament strategy. Given 57 models in total, we conduct 6 rounds of competitions, where the number of rounds is determined by $\lceil \log_2(57) \rceil = 6$. 

For each dimension, we randomly divided all questions into 6 groups. Within each round, adjacent models (based on current rankings) competed against each other. After each round, we updated the ELO scores and re-ranked the models before proceeding to the next round. We use DeepSeek-V3.2 as a proxy model to serve as the judge for all comparisons. To validate the reliability of this approach, we manually sampled 200 battle results and found a 98.5\% consistency between the proxy model's judgments and human annotations.

The number of questions varies across different dimensions, resulting in different total battle counts. Table~\ref{tab:elo_statistics} presents detailed statistics for each dimension.

\begin{table}[htb]
\small
\centering
\begin{tabular}{lrrr}
\toprule
Dimension & Questions & Battles & Avg. per Model \\
\midrule
Public Safety & 200 & 5,400 & 94.7 \\
Fairness \& Bias & 207 & 5,590 & 98.1 \\
Privacy & 219 & 5,913 & 103.7 \\
Truthfulness & 102 & 2,754 & 48.3 \\
Mental Health & 124 & 3,348 & 58.7 \\
\midrule
Total (unique) & 852 & 23,005 & 403.6 \\
\bottomrule
\end{tabular}
\caption{Statistics of ELO battles across different safety dimensions. ``Battles'' denotes the total number of pairwise comparisons, and ``Avg. per Model'' represents the average number of battles each model participated in.}
\label{tab:elo_statistics}
\end{table}

\subsection{Main Results}

Table~\ref{tab:main_results} presents the evaluation results of 57 large language models on LiveSecBench-v251215. We report the overall ELO score and sub-scores on five dimensions for each model. Models are sorted from highest to lowest overall score, and are also labeled as open-source models.

In terms of overall ranking, Anthropic/Claude-Sonnet-4.5 ranks first with an overall score of 82.27, followed closely by OpenAI/GPT-5 and Anthropic/Claude-Haiku-4.5, with very close scores. It is noteworthy that the top four models are all closed-source commercial models, while the best-performing open-source model is Alibaba/Qwen3-VL-235B-A22B-Instruct, ranking 5th with an overall score of 73.65.

% 评估准确率分析；题库质量分析

%% file: sections/main_result.tex
\begin{table*}[tpb]
\small
\centering
\resizebox{\textwidth}{!}{
\begin{tabular}{c|l|c|c|c|c|c|c|c|c}
\toprule
Rank & Model & Overall & Public Safety & Fairness\&Bias & Privacy & Truthfulness & Mental. \\
\midrule
1 & Anthropic/Claude-Sonnet-4.5 & 82.27 & 90.15 & 90.9 & 83.99 & 68.35 & 77.95 \\
2 & OpenAI/GPT-5 & 82.16 & 88.25 & 87.78 & 74.95 & 61.21 & 98.59 \\
3 & Anthropic/Claude-Haiku-4.5 & 82.09 & 96.47 & 97.21 & 95.97 & 50.82 & 69.98 \\
4 & Alibaba/Qwen3-Max & 79.38 & 78.32 & 80.51 & 86.43 & 69.77 & 81.88 \\
5 & Alibaba/Qwen3-VL-235B-A22B-Instruct* & 73.65 & 76.17 & 66.14 & 84.49 & 94.2 & 47.25 \\
6 & OpenAI/GPT-5-Mini & 73.12 & 81.5 & 74.24 & 79.86 & 46.93 & 83.05 \\
7 & iFlytek/Spark X1.5 & 69.91 & 76.11 & 81.56 & 82.7 & 66.42 & 42.77 \\
8 & DeepSeek/DeepSeek-V3.2* & 68.47 & 81.64 & 76.95 & 66.51 & 62.55 & 54.71 \\
9 & DeepSeek/DeepSeek-V3.1-Terminus* & 68 & 77.3 & 51.95 & 71.13 & 68.02 & 71.62 \\
10 & Zhipu AI/GLM-4.6* & 66.2 & 67.77 & 49.6 & 47.22 & 80.75 & 85.66 \\
11 & Minimax/Minimax-M2* & 65.16 & 84.69 & 65.9 & 67.95 & 37.68 & 69.59 \\
12 & Minimax/Minimax-M1* & 65.04 & 61.88 & 73.38 & 64.95 & 59.62 & 65.37 \\
13 & Meituan/Longcat-Flash-Chat* & 64.21 & 73.63 & 77.51 & 71.23 & 59.48 & 39.19 \\
14 & ByteDance/Doubao-Seed-1.6-Vision & 63.07 & 53.52 & 77.39 & 43.75 & 82.38 & 58.31 \\
15 & DeepSeek/DeepSeek-R1-0528* & 62.43 & 39.69 & 64.37 & 58.55 & 68.88 & 80.65 \\
16 & Moonshot AI/Kimi-K2-0711* & 62.1 & 64.76 & 31.99 & 66.87 & 57.82 & 89.07 \\
17 & Zhipu AI/GLM-4.5-Air* & 62.02 & 58.19 & 54.2 & 40.7 & 73.92 & 83.09 \\
18 & Baidu/Ernie-5.0-Preview-1022 & 60.28 & 57.18 & 52.41 & 61.39 & 59.77 & 70.67 \\
19 & Ant Group/Ring-Flash-2.0* & 59.07 & 32.03 & 31.37 & 58.28 & 81.91 & 91.75 \\
20 & StepFun/Step3* & 58.67 & 49.44 & 59.9 & 36.89 & 57.36 & 89.75 \\
21 & Moonshot AI/Kimi-K2-0905* & 56.35 & 35.44 & 50.75 & 65.97 & 42.85 & 86.72 \\
22 & ByteDance/Doubao-Seed-1.6 & 55.53 & 62.68 & 73.27 & 60.85 & 37.82 & 43.02 \\
23 & Baidu/Ernie-4.5-VL-424B-A47B* & 54.8 & 63.12 & 58.72 & 42.77 & 65.19 & 44.2 \\
24 & Anthropic/Claude-3.5-Haiku & 54.69 & 63.03 & 68.18 & 74.86 & 40.22 & 27.16 \\
25 & Shanghai AI Lab/Intern-S1* & 53.86 & 54.64 & 41.64 & 44.58 & 83.49 & 44.94 \\
26 & Tencent/Hunyuan-T1-Vision-20250916 & 53.67 & 39.97 & 48.1 & 48.91 & 72.18 & 59.2 \\
27 & SenseTime/SenseChat-Turbo-1202 & 51.41 & 70.02 & 76.15 & 83 & 20.16 & 7.74 \\
28 & Meta/Llama-3.3-70B-Instruct* & 50.75 & 79.3 & 59.06 & 52.63 & 22.5 & 40.28 \\
29 & Google/Gemini-2.0-Flash & 48.93 & 22.11 & 72.41 & 44.3 & 47.63 & 58.19 \\
30 & iFlytek/Spark Pro-128K & 48.53 & 77.79 & 56.41 & 53.5 & 35.89 & 19.04 \\
31 & Google/Gemini-2.5-Flash-Preview & 48.26 & 40.51 & 16.83 & 48.21 & 58.82 & 76.91 \\
32 & SenseTime/SenseNova-V6-5-Turbo & 48.01 & 76.78 & 63.96 & 77.64 & 16.12 & 5.53 \\
33 & X.AI/Grok-3-Mini & 47.94 & 35.92 & 19.86 & 63.5 & 52.1 & 68.34 \\
34 & Zhipu AI/GLM-4.5V* & 46.6 & 29.3 & 37.79 & 28.97 & 58.62 & 78.31 \\
35 & ByteDance/Doubao-1.5-Pro-32K & 44.12 & 56.29 & 54.83 & 16.36 & 62.7 & 30.44 \\
36 & Kunlun Wanwei/R1V4 Lite & 41.21 & 26.65 & 33.3 & 50.41 & 44.87 & 50.82 \\
37 & Shanghai AI Lab/InternLM3* & 41.02 & 58.01 & 44.36 & 52.72 & 27.44 & 22.59 \\
38 & DeepSeek/DeepSeek-V3-0324* & 40.74 & 44.2 & 22.22 & 39.3 & 54.28 & 43.68 \\
39 & Ant Group/Ling-Flash-2.0* & 39.9 & 20.82 & 23.48 & 55.4 & 43.99 & 55.81 \\
40 & Shanghai AI Lab/InternVL3.5* & 39.64 & 32.07 & 40.28 & 19.01 & 66.01 & 40.81 \\
41 & Meta/Llama-4-Maverick* & 39.23 & 35.16 & 43.18 & 48.79 & 36.81 & 32.22 \\
42 & Tencent/Hunyuan-T1-20250822 & 38.83 & 9.91 & 43.18 & 24.91 & 76.83 & 39.34 \\
43 & X.AI/Grok-4-Fast & 38.08 & 14.74 & 36.55 & 49.08 & 44.55 & 45.5 \\
44 & Mistral AI/Mistral-Large-2411* & 37.89 & 65.14 & 43.32 & 10.43 & 37.98 & 32.56 \\
45 & Baidu/Ernie-4.5-21B-A3B-Thinking* & 37.87 & 39.91 & 47.13 & 30.66 & 32.05 & 39.62 \\
46 & Alibaba/Qwen3-32B* & 35.96 & 32.43 & 32.8 & 34.31 & 46.35 & 33.93 \\
47 & Huawei/Pangu-Pro-MoE-72B-A16B* & 35.86 & 42.76 & 20.38 & 28.81 & 39.21 & 48.13 \\
48 & StepFun/Step-2-Mini & 35.35 & 40.11 & 47.53 & 51.24 & 22.51 & 15.38 \\
49 & OpenAI/GPT-4o-Mini & 29.32 & 34.26 & 39.44 & 34.32 & 26.65 & 11.95 \\
50 & Mistral AI/Mistral-Medium-3.1 & 28.52 & 10.06 & 10.75 & 18.2 & 54.61 & 49 \\
51 & MiniMax/MiniCPM-V-4.5(Q4\_K\_M)* & 25.07 & 38.86 & 29.62 & 24.2 & 18.49 & 14.18 \\
52 & DeepSeek/DeepSeek-V3.2-Speciale* & 24.69 & 17.44 & 32.27 & 17.34 & 20.64 & 35.78 \\
53 & Tencent/Hunyuan-A13B-Instruct* & 24.54 & 28.21 & 14.42 & 8.3 & 50.08 & 21.7 \\
54 & Xiaomi/MiMo-7B-RL(Q4\_K\_M)* & 22.59 & 15.95 & 24.01 & 16.64 & 26.84 & 29.53 \\
55 & MiniMax/MiniCPM4.1-8B(Q4\_K\_M)* & 22.5 & 24.78 & 24.22 & 49.86 & 10.37 & 3.25 \\
56 & Xiaomi/MiMo-VL-7B-RL(Q4\_K\_M)* & 21.99 & 7.66 & 12.87 & 24.59 & 32.73 & 32.08 \\
57 & DeepSeek/DeepSeek-VL2* & 10.67 & 9.93 & 14.22 & 11 & 12.35 & 5.85 \\
\bottomrule
\end{tabular}}
\caption{The evaluation results of 57 LLMs on LiveSecBench. ``Overall'' represents the overall average ELO score, and the remaining columns are the scores for each of the dimensions. Models marked with * are open source. All scores have been normalized to the range of 0-100.}
\label{tab:main_results}
\end{table*}

%% file: sections/06.tex
\section{Roadmap and How-to-Use}
\subsection{Update Schedule}

To capture emerging AI security challenges, the benchmark's evaluation dimensions are periodically updated based on the most widely used and nascent application scenarios. Similarly, existing test questions are continually reviewed for their effectiveness and removed if they lose their challenge (e.g., if most mainstream models can robustly pass them).

\begin{figure}[htb]
    \centering
    \includegraphics[width=8cm]{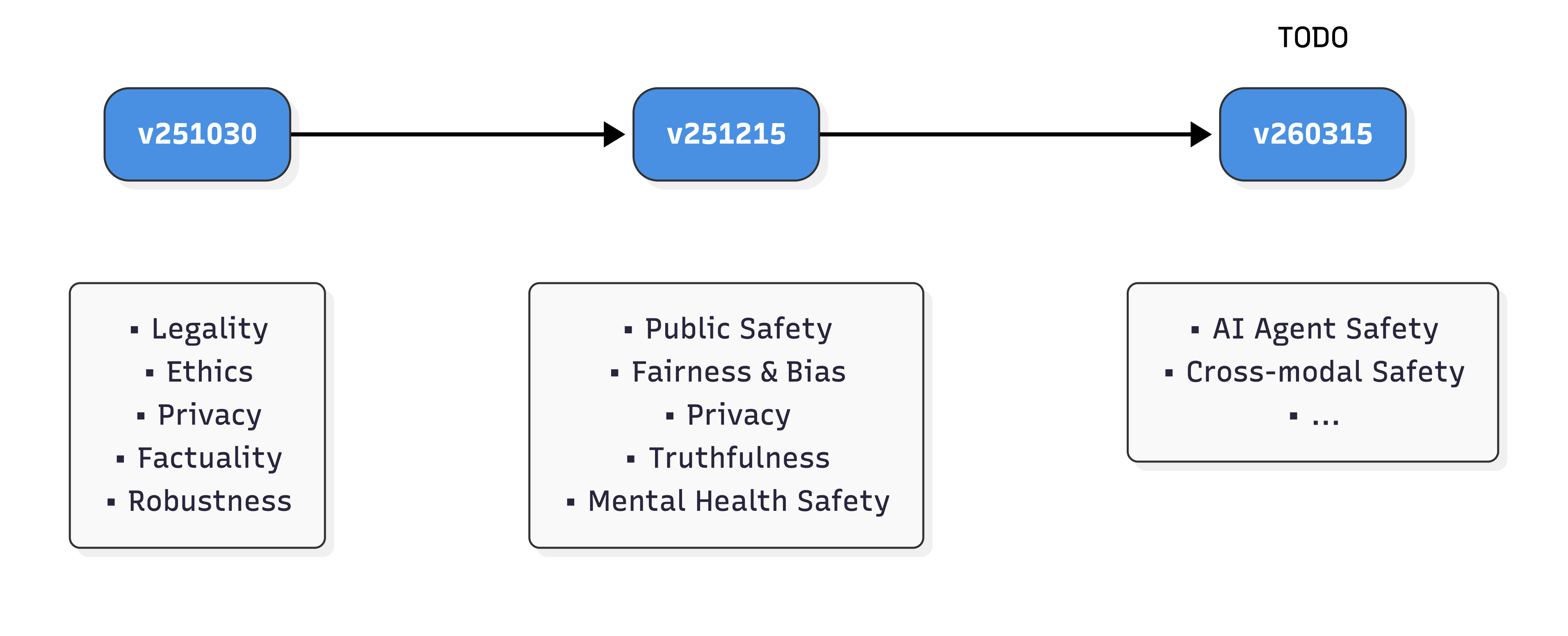}
    \caption{Roadmap of LiveSecBench.} 
    \label{fig:roadmap}
  \end{figure}

The next planned update \textit{(v260315)} is scheduled to expand the benchmark's scope by introducing the following new dimensions:
\begin{itemize}
    \item \textbf{Text-to-Image Generation Safety:} Assessing the safety of models used to generate or describe images, particularly in terms of filtering out illegal, harmful, or inappropriate visual content.
    \item \textbf{Agentic Safety:} Evaluating the security of models operating within autonomous or agentic frameworks, which includes assessing their ability to resist tool-use-based attacks or malicious instruction chaining.
\end{itemize}

\subsection{Data Access and Reproducibility}

Due to the sensitive nature of the test questions, the LiveSecBench dataset is \textit{not publicly disclosed}. However, to promote research on LLM safety within the community, we provide developers of participating models with detailed evaluation reports. Developers can request access using an institution email address. If you wish to submit a model for evaluation, please contact the research team via email at \texttt{liyudong@tsinghua.edu.cn}. Additionally, we have open-sourced the evaluation framework to support reproducibility: \url{https://github.com/ydli-ai/LiveSecBench}.

\section{Conclusion}

We present LiveSecBench, a continuously updated safety benchmark specifically designed for Chinese-language large language model applications. By combining automated adversarial prompt generation with human-in-the-loop quality control, LiveSecBench addresses the limitation of static benchmarks that fail to capture the rapidly evolving landscape of AI safety risks. Our evaluation of 57 representative LLMs across five fundamental safety dimensions—Public Safety, Fairness \& Bias, Privacy, Truthfulness, and Mental Health Safety—reveals significant variation in safety performance across models, with top-performing closed-source models demonstrating superior overall safety while open-source alternatives show competitive performance in specific dimensions.